\documentclass[letterpaper]{article} % DO NOT CHANGE THIS
\usepackage{aaai24}  % DO NOT CHANGE THIS
\usepackage{times}  % DO NOT CHANGE THIS
\usepackage{helvet}  % DO NOT CHANGE THIS
\usepackage{courier}  % DO NOT CHANGE THIS
\usepackage[hyphens]{url}  % DO NOT CHANGE THIS
\usepackage{graphicx} % DO NOT CHANGE THIS
\usepackage{natbib}  % DO NOT CHANGE THIS AND DO NOT ADD ANY OPTIONS TO IT
\usepackage{caption} % DO NOT CHANGE THIS AND DO NOT ADD ANY OPTIONS TO IT

\usepackage{algorithm}
\usepackage{algorithmic}

\usepackage{epsfig}
\usepackage{amsmath}
\usepackage{amssymb}
\usepackage{booktabs}
\usepackage{subfigure}
\usepackage{multirow}
\usepackage{pifont}

\usepackage{newfloat}
\usepackage{listings}
\DeclareCaptionStyle{ruled}{labelfont=normalfont,labelsep=colon,strut=off} % DO NOT CHANGE THIS
\floatstyle{ruled}
\newfloat{listing}{tb}{lst}{}
\floatname{listing}{Listing}
\title{Unlocking the Power of Open Set: A New Perspective for Open-Set Noisy Label Learning}
\author{
    Wenhai Wan\equalcontrib, Xinrui Wang\equalcontrib, Ming-Kun Xie, Shao-Yuan Li\thanks{Corresponding author.}, Sheng-Jun Huang, Songcan Chen
}
\affiliations{
    \textsuperscript{}College of Computer Science and Technology, Nanjing University of Aeronautics and Astronautics\\
    MIIT Key Laboratory of Pattern Analysis and Machine Intelligence
}

\usepackage{bibentry}
\begin{document}

\maketitle

\begin{abstract}
Learning from noisy data has attracted much attention, where most methods focus on closed-set label noise. However, a more common scenario in the real world is the presence of both open-set and closed-set noise. Existing methods typically identify and handle these two types of label noise separately by designing a specific strategy for each type. However, in many real-world scenarios, it would be challenging to identify open-set examples, especially when the dataset has been severely corrupted. Unlike the previous works, we explore how models behave when faced with open-set examples, and find that \emph{a part of open-set examples gradually get integrated into certain known classes}, which is beneficial for the separation among known classes. Motivated by the phenomenon, we propose a novel two-step contrastive learning method CECL (Class Expansion Contrastive Learning) which aims to deal with both types of label noise by exploiting the useful information of open-set examples. Specifically, we incorporate some open-set examples into closed-set classes to enhance performance while treating others as delimiters to improve representative ability. Extensive experiments on synthetic and real-world datasets with diverse label noise demonstrate the effectiveness of CECL.
\end{abstract}

\section{Introduction}
\label{submission}

Deep neural networks (DNNs) have achieved remarkable success in various tasks. The great success is primarily attributed to large amounts of data with high-quality annotations, which are expensive or even inaccessible in practice. Actually, datasets collected via search engines or crowdsourcing platforms inevitably involve noisy labels \cite{xiao2015learning,li2018multi,li2021crowdsourcing,shi2023learning}. Given the powerful learning capacities of DNNs, the model will ultimately overfit the label noise and lead to poor generalization performance \cite{pmlr-v70-arpit17a, zhang2021understanding}. To mitigate this issue, it is significant to develop robust models for learning from noisy labels.

\begin{figure}[t]
% \vskip 0.2in
\begin{center}
\centerline{\includegraphics[width=\columnwidth]{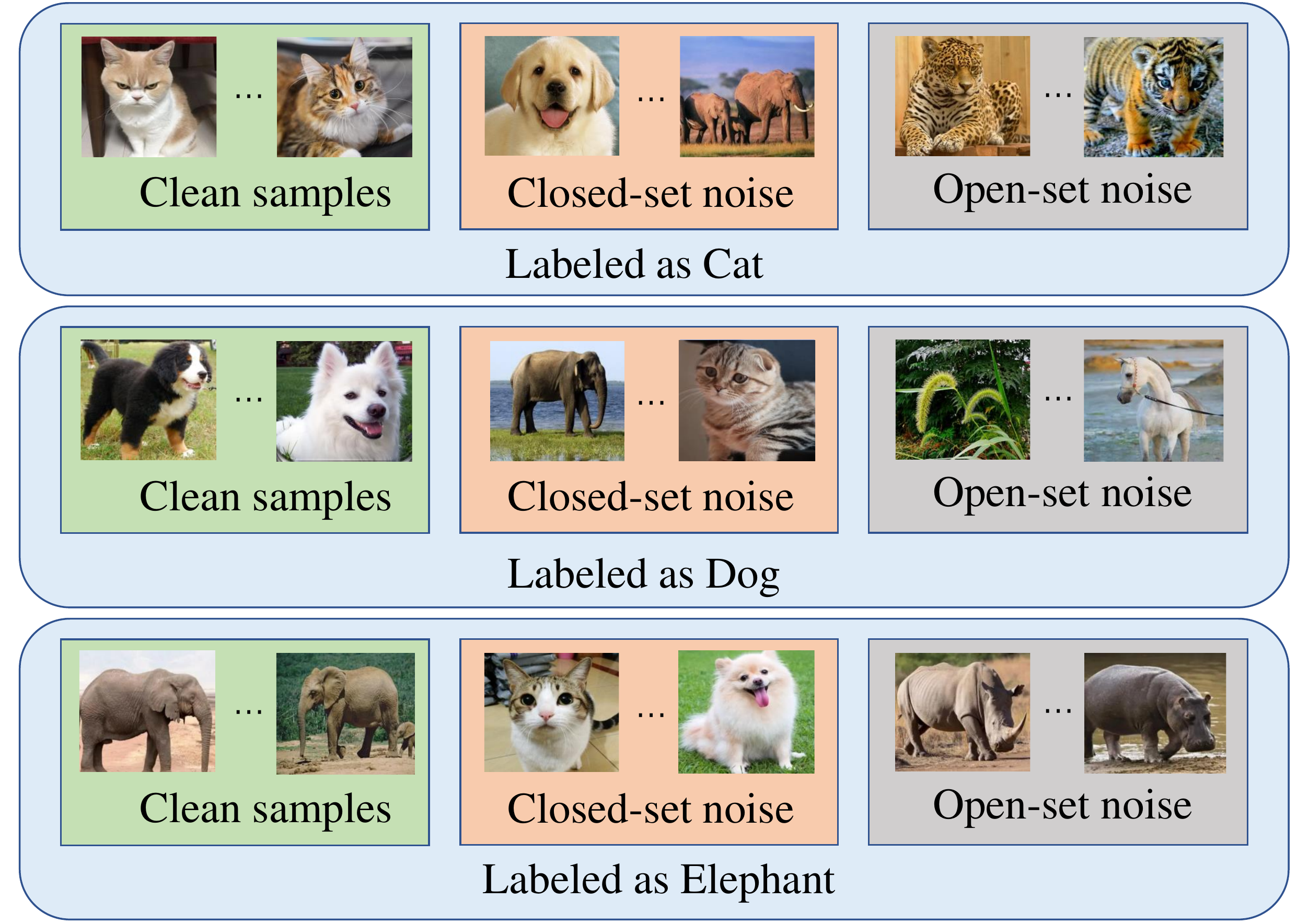}}
\caption{An example of open-set noisy label learning problem. $\emph{\{cat, dog, elephant\}}$ is the concerned \emph{known classes}. The left, middle, and right columns respectively show images that are correctly labeled, wrongly labeled with \emph{closed-set} and \emph{open-set} noise.
}
\label{problem-setting}
\end{center}
\end{figure}

As shown in Figure \ref{problem-setting}, we divide the mislabeled examples into two types: \emph{closed-set} and \emph{open-set}. More specifically, a \emph{closed-set} mislabeled example occurs when its true class fall within the set of known classes $\emph\{cat, dog, elephant\}$, while an \emph{open-set} mislabeled example occurs when its true class does not fall within the set of known classes in the training data. To our knowledge, existing works mainly focus on closed-set scenarios \cite{li2020dividemix, li2022selective}, while a more common scenario in the real world is the presence of both \emph{closed-set} and \emph{open-set}.

%\cite{wang2018iterative} proposed an iterative learning framework (pcLOF) to detect OSN, and then take a re-weighting strategy in objective function; \cite{yao2021jo} proposed an approach to detect OSN based on consistency, and then ignore these OSN; \cite{sun2022pnp} adopted an OSN detection approach based on consistency, and then maximized the entropy on these detected OSN. To sum up, these strategies can be roughly divided into three groups: \textbf{Down-Weighting}, \textbf{Discarding}, and \textbf{Entropy Maximization} %Furthermore, we surveyed common ways of treating OOD in the area of out-of-distribution\cite{chan2021entropy, geng2020recent, zhou2022model}. 
%The most challenging issue in the above OSNLL (Open-Set Noisy Label Learning) problem arises due to the presence of open-set examples. 
The problem has been formalized as a learning framework called Open-Set Noisy Label Learning (OSNLL)  \cite{wang2018iterative}, which is also the main focus of our work.
The most challenging issue in OSNLL arises due to the presence of open-set examples. The detection and handling of these examples pose a significant challenge, especially when their distribution is unknown. Several papers  \cite{wang2018iterative,yao2021jo,sun2022pnp} have proposed specific methods and frameworks by identifying mislabeled open-set examples and minimizing their impact. However, empirical findings \cite{morteza2022provable} and theoretical results \cite{fang2022out} suggest that some open-set examples become increasingly difficult for the model to distinguish. In some cases, it is even impossible to recognize open-set examples from clean datasets, making learning with both closed-set and open-set examples particularly challenging.

In this paper, we give a more in-depth study of the effect of open-set examples on OSNLL. Specifically, we first investigate the model's training behavior when facing open-set examples, observing that some open-set classes get integrated with closed-set classes. By treating these open-set examples as closed-set ones, the models surprisingly acquire substantial enhancement in the classification ability. We call this the \textbf{Class Expansion} phenomenon. An intuitive explanation is, by introducing these "false positive" examples from the unknown class data, the representation of different known classes becomes more generalized.

Based on the above observations, we propose a novel two-step Class Expansion Contrastive Learning (CECL) framework to make full use of open-set examples while minimizing the negative impact of label noise. In the first step, we roughly tackle the data noise and maintain the basic concept of clean and noisy examples. In the second step, we adopt a contrastive learning scheme and selectively incorporate certain indistinct open-set examples into their corresponding similar known classes. In fact, the concept of known classes is broadened to include these similar, but previously unknown examples. Additionally, to enhance the discrimination between different classes, we use the remaining distinguishable open-set examples as delimiters.
 
%It is worth noting that we follow the consistent setting with previous work\cite{wang2018iterative, yao2021jo, sun2022pnp}, in which open-set examples only appear during the training phase, but not exist during the testing phase. We do not need to consider open-set recognition.
% which is also the main focus of our work, it is worth noting that in this setting, we only have open-set examples during the training phase, while they do not exist during the testing phase, which is consistent with previous work\cite{wang2018iterative, yao2021jo, sun2022pnp}. In other words, we do not need to consider open-set recognition.

In summary, our contributions are threefold:
\begin{itemize}

\item We explore the relationship between open-set and closed-set classes and highlight the phenomenon of Class Expansion in OSNLL, demonstrating the potential of open-set examples to facilitate known class learning. %Our paper is the first one to leverage these properties in the fields of OSNLL.

\item We propose a novel CECL framework that incorporates contrastive learning to fully utilize open-set examples, which better generalizes and discriminates the representation of different known classes. 

\item
  We provide theoretical guarantees for better representation capability of CECL, and conduct extensive experiments on both synthetic and real-world noisy datasets to validate its effectiveness.
  \end{itemize}

%------------------------------------------------------------------------

\section{Related Work}
\textbf{Noisy Label Learning} Most of the methods in literature mitigate the label noise by robust loss functions\cite{wang2019symmetric,wang2019imae,xu2019l_dmi,liu2020early,wei2021open}, noise transition matrix\cite{patrini2017making,tanno2019learning,xia2019anchor,li2022estimating}, sample selection\cite{han2018co,yu2019does,wei2020combating,li2022selective,huang2023twin,xia2023combating}, and label correction\cite{li2020dividemix,li2021learning,ortego2021multi,zhang2023rankmatch}. 

\noindent\textbf{Open-set Noisy Label Learning} Open-set examples in the real-world are ubiquitous\cite{YangWSLZXY22,PAL}. Open-set noisy label learning aims to develop a well-performing model from real-world datasets that contain both closed-set and open-set noisy labels. \cite{wang2018iterative} proposes an iterative learning framework called pcLOF to detect open-set noisy labels and then uses a re-weighting strategy in the objective function. \cite{yao2021jo,sun2022pnp} introduces an approach based on consistency to identify open-set examples, and then minimize the impact they bring, \cite{li2020mopro} proposes momentum prototypes for webly-supervised representation learning which lacks consideration for the diversity of open-set examples, \cite{wu2021ngc} introduces a noisy graph cleaning framework that simultaneously performs noise correction and clean data selection, \cite{xia2022extended} designs an extended $T$-estimator to estimate the cluster-dependent transition matrix by exploiting the noisy data. These methods rest on the assumption that the open-set examples are harmful to known class learning, which is questionable as we will show in this work.

%------------------------------------------------------------------------

%\section{Methodology}\label{Method}
%In this section, we will begin by providing a comprehensive explanation of the proposed class expansion phenomenon in Section \ref{clsexp} and then describe how to integrate this phenomenon into OSNLL. Specifically, our framework consists of two essential steps that work together to unlock the potential of open-set examples. Both steps are crucial and cannot be overlooked. More detailed information can be found in Section \ref{cecl}.  -----(此段缩减篇幅)

\section{The Class Expansion Phenomenon} \label{clsexp}
In real-world scenarios, open-set examples are pervasive, posing significant challenges to many problems. Many researchers strive to distinguish open-set examples and mitigate their influence. In fact, each inclusion of an open-set example in the dataset has a reason behind it, either due to misjudgment, inherent attributes that are easily confused, or a lack of clear definition of the open-set, and so on. It might be unreasonable to simply treat all open-set examples caused by various reasons as a single type of outlier! Furthermore, some open-set examples may share similar features with closed-set examples, which can be confusing even for humans. Simply eliminating the impact of open-set examples may potentially exert negative effects on the learning of closed-set examples. So, is there room for us to explore a different avenue? 
\begin{figure}[htbp]
\begin{center}
\centerline{\includegraphics[width=\columnwidth]{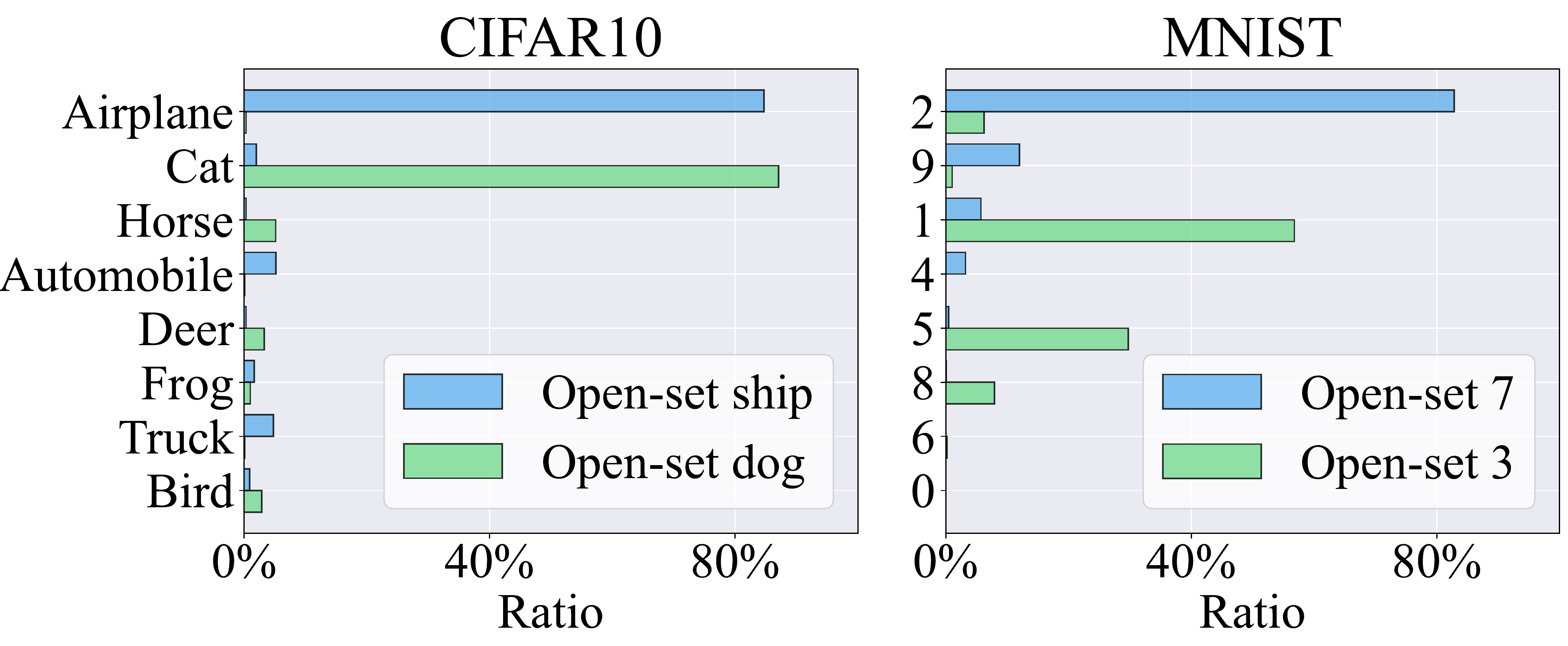}}
\caption{The distribution of open-set examples among different known classes on CIFAR10 and MNIST, respectively.}
\label{valid}
\end{center}
\end{figure}

In the pursuit of an answer to the above question, we investigate the behavior of DNNs when encountering unseen open-set examples. Specifically, we conducted a series of experiments on the CIFAR10 \cite{krizhevsky2009learning}, MNIST \cite{lecun1998mnist}  and Tiny-Imagenet \cite{le2015tiny} datasets.
We randomly select $80\%$ of the total categories as known classes and treat the remaining classes as unknown for all datasets. We train the model on labeled closed-set examples and generated pseudo-labels on open-set examples. Figure \ref{valid} illustrates the class distribution of pseudo-labels generated by the model. On the CIFAR10 dataset, it can be observed that open-set \textit{dog} examples are almost uniformly classified as \textit{cat}. This phenomenon occurs mainly due to the fact that \textit{dog} and \textit{cat} share high-level similarity, making the model hard to distinguish between these two semantic objects and merge them into a generalized concept of \textit{cat}. We use the term "\textbf{class expansion}" to describe this phenomenon. The class expansion phenomenon can also be found on MNIST (e.g. open-set digit \textit{7} are consistently being classified as closed-set class \textit{2}), however, it is noteworthy that not all open-set class concepts can be treated as class expansion to a specific closed-set class concept (e.g. open-set digit \textit{3} are relatively evenly distributed among multiple closed-set classes). Our observations indicate that when encountering open-set examples, the model tends to make reasoned judgments rather than reckless ones. \footnote{A similar phenomenon is observed on Tiny-Imagenet, and we provide the transition matrix in the Appendix.}

\begin{figure}[ht]
	\centering
  \subfigure[Test accuracy comparison.]{
\includegraphics[width=0.8\columnwidth,height=1in]{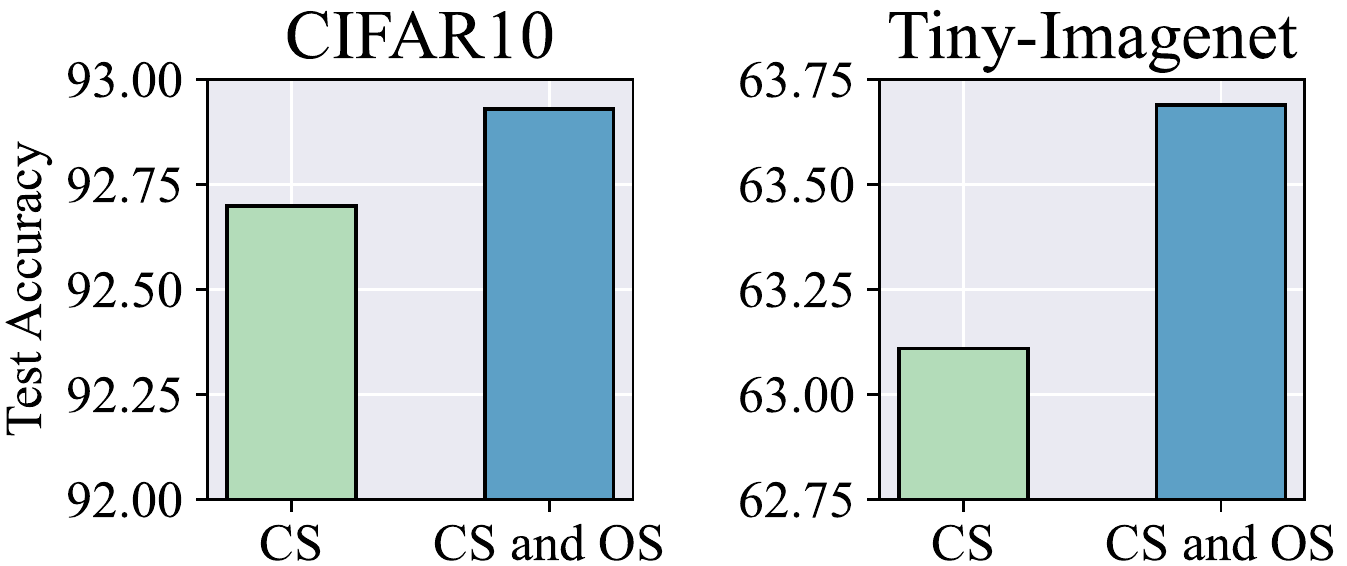}}

	 \subfigure[Training process on Tiny-Imagenet.]{
\includegraphics[width=0.78\columnwidth,height=1in]{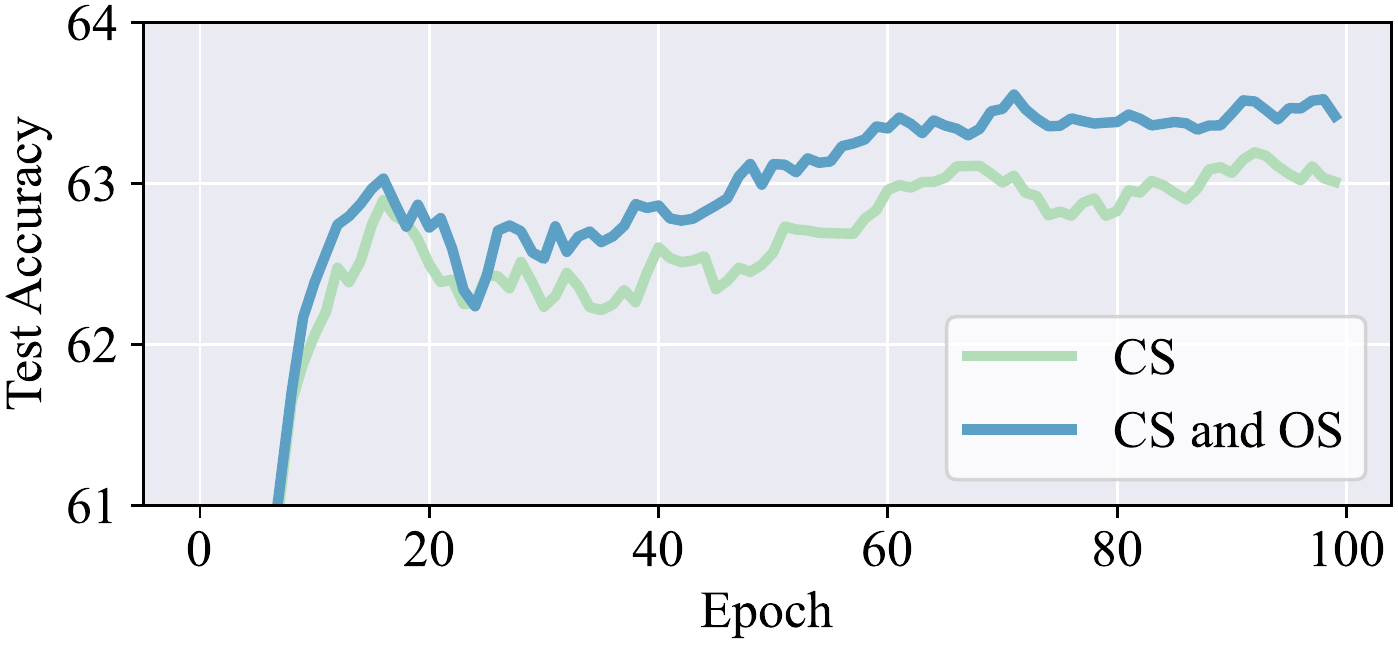}	}
	 
	\caption{Experimental results of incorporating examples from open-set classes into known classes for learning on CIFAR10 and Tiny-Imagenet. 'CS', 'CS and OS' respectively denote training only on closed-set class examples and a mixture of closed-set and open-set class examples. }\label{fig:ill_exp}
\end{figure}

% \begin{table}
% \centering
% \caption{Experimental results of incorporating examples from open-set classes into known classes for learning on CIFAR10, Tiny-Imagenet separately, 'CS' denotes training only on closed-set class examples, while 'CS and OS' indicate training with a mixture of closed-set and open-set class examples.}
% \begin{tabular}{c|c|c} 
% \toprule
%                        & CIFAR10 & Tiny-Imagenet  \\  \hline 
% CS                  &    92.70    &       63.11             \\               
% CS and OS    &    92.93    &       63.69            \\
% \bottomrule

% \end{tabular}

% \end{table}

%With the above observations, a natural question arises: can we incorporate open-set examples to aid in learning closed-set classes? To investigate this, 

To further explore the effectiveness of class expansion, we perform simple experiments by selecting those open-set examples with high prediction scores on closed-set classes into the training set and retraining the model. % Our exploration is conducted on two datasets, CIFAR10 and Tiny-Imagenet \cite{le2015tiny}. Specifically, in both datasets, we randomly select $20\%$ of the total categories as open-set classes. 
 Figure \ref{fig:ill_exp} (a) presents the results obtained by training using only closed-set examples (CS) and training using both closed-set examples and open-set examples simultaneously (CS and OS). It is surprising that the incorporation of open-set examples significantly improves the model's performance on closed-set classes, which contradicts our common sense. It is also worth noting that the improvement becomes more obvious as the dataset's scale increases, see the training process for Tiny-Imagenet shown in Figure \ref{fig:ill_exp} (b). The above findings indicate the potential power of open-set examples to boost known class recognition. Below, we amplify the usefulness of open-set examples by incorporating the contrastive learning into model training, leading to our Class Expansion Contrastive Learning (CECL) method.

\section{Methodology }\label{cecl}

Let $\mathcal{X}$ denotes the instance space and $\mathcal{Y} = \{1, 2, ..., c\}$ the label space with $c$ distinct classes. We use $\mathcal{D} = \{ (x_i, y_i)|1 \le i \le n \}$ to denote the training dataset, with $y_i \in [0, 1]^c$ denoting the one-hot label vector over $c$ classes. 
For datasets with noisy labels,  $\{y_i\}$ are not guaranteed to be correct. Formally, we represent the ground-truth label of example $x_i$ with a one-hot label vector $\hat{y}_i\in [0,1]^{c+1}$  over $c+1$ classes, where the $(c+1)$-th class indicates the open-set class. It is noteworthy that $\hat{y}_i$ is unknown throughout the entire training process. Due to the presence of closed-set and open-set label noise, directly training a model on $\mathcal{D}$ fails to achieve favorable generalization performance.

\begin{figure}[ht!]
    \centering
    \includegraphics[width=0.8\columnwidth]{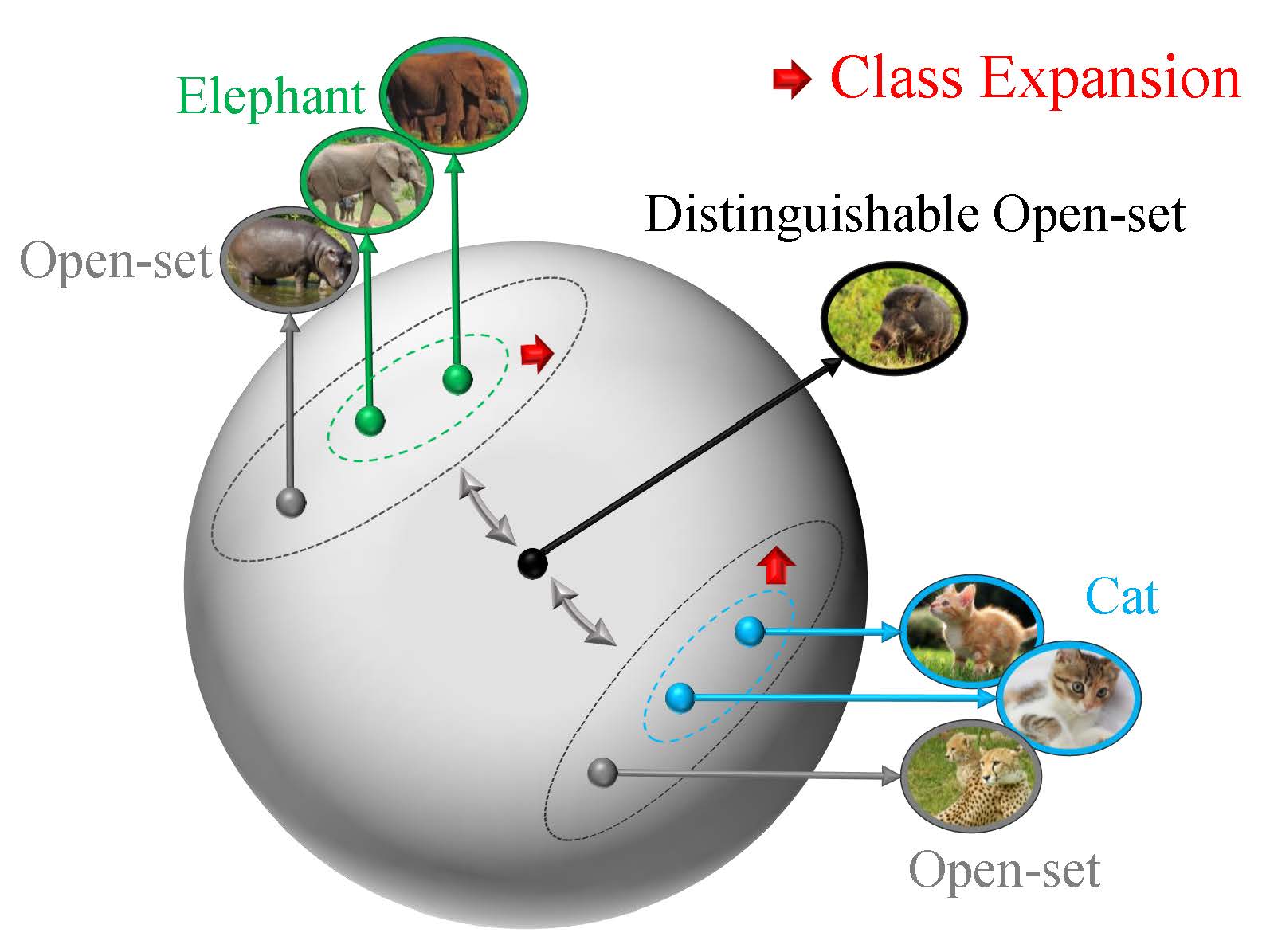}
    \caption{The intuition of CECL. CECL incorporates certain indistinct open-set examples into the known classes, which are expected to contribute to class expansion with better generalization. Additionally, the distinguishable open-set examples are used as delimiters, which are expected to push away between the known classes with better discrimination.}
    \label{delimiter}
\end{figure}

In this paper, building on the comprehensive success of contrastive learning in extensive tasks, we propose a novel class expansion contrastive learning (CECL) framework to better leverage and unlock the power of open-set examples. Figure \ref{delimiter} shows the intuition of CECL. Provided that we have detected representative examples for the known classes, then the \emph{distinguishable} open-set examples that possess very different features from the known class examples can be reliably detected. The remaining examples can be either known class examples or class expansion contributing open-set examples. From the contrastive learning perspective, we can wisely use these two types of open-set examples to push away the classes and better generalize their concept boundary. In the following, we introduce the details.

\begin{figure*}[htbp]
    \centering
    \includegraphics[width=0.8\textwidth]{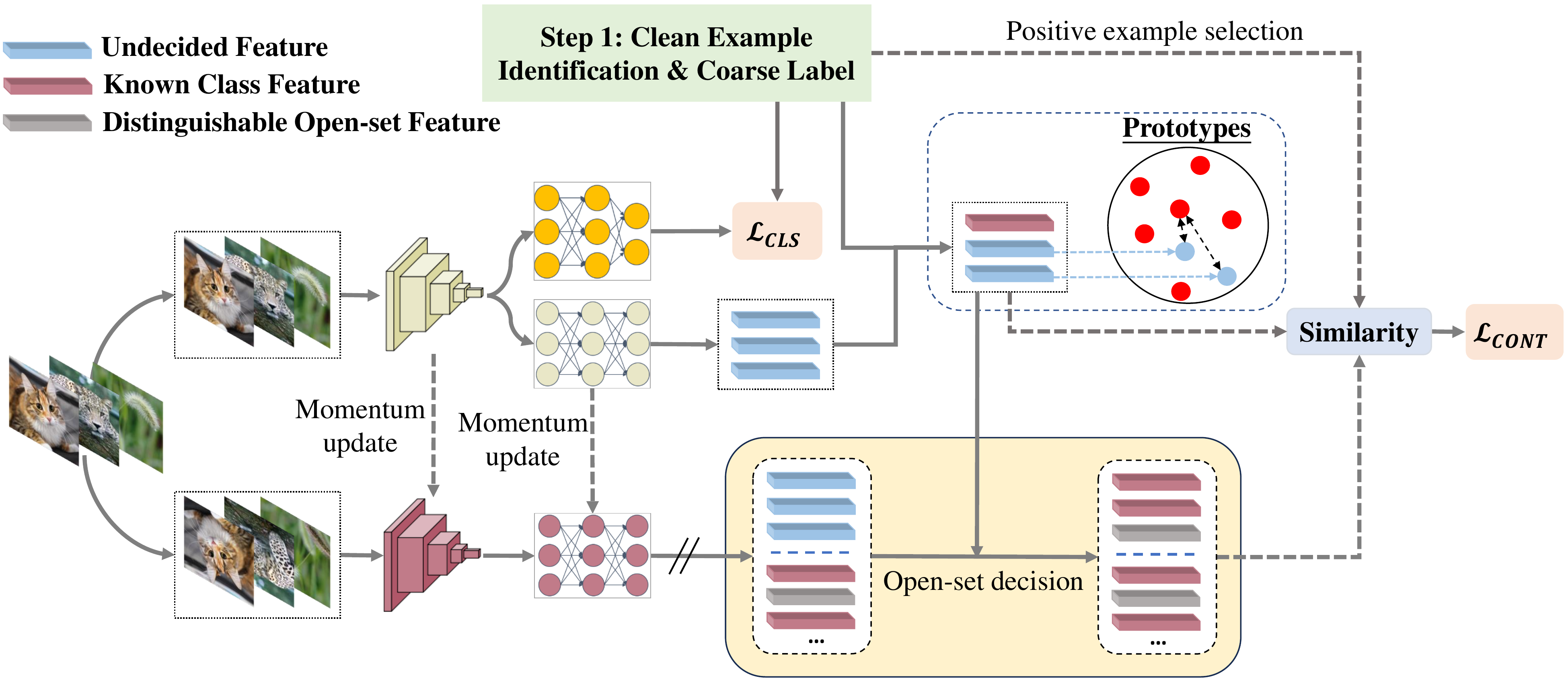} 
        \caption{Illustration of CECL. According to the information obtained in the first step, clean examples are used to generate prototypes for each class, certain open-set examples are incorporated into known classes in the form of class expansion, and remaining are perceived as delimiters. The momentum embeddings are maintained by a queue structure. '//' means stop gradient.}
    \label{frame}      
\end{figure*}
\subsection{Class Expansion Contrastive Learning}
Although supervised contrastive learning has been extensively studied, its application in OSNLL has not been explored due to two main challenges. First, it is difficult to construct the positive example set, second, it is challenging to determine whether an open-set example should be included in the positive example set. 

To solve the above challenges, we introduce an efficient and robust two-step framework.
In the first step, we perform a pretraining phase to obtain coarse labels for the entire dataset
and identify whether each example is clean or not.

In the second step, we leverage the predictions from the first step to generate cluster prototypes for each class, which we then use to perform label noise filter and open-set decision. The identified distinguishable open-set examples are used to enhance the model’s representation learning ability in our improved contrastive learning framework. Overall, our proposed framework provides an effective approach for training models on datasets with noisy labels in open-set scenarios. Our approach is illustrated in Figure \ref{frame}.

\noindent\textbf{Step 1: Clean Example Identification.} To ensure more efficient training in the subsequent step, we conduct the first step in our proposed method. Specifically, we adopted one effective off-the-shelf closed-set noisy label learning method Promix \cite{wang2022promix} to identify a subset of clean examples. Promix adopts a dual network co-teaching
learning style, we further ameliorate it with one correction record $T$ indicating whether one instance is mislabeled. 
$T(\cdot)$ is a $0/1$ valued indicator that records whether the model's prediction is the same as the instance's label during the whole training process. Then, we can divide the original noisy dataset $\mathcal{D}$ into two subsets $\mathcal{D}_{clean}$ and $\mathcal{D}_{noisy}$. $\mathcal{D}_{clean}$ includes the examples that are highly believed to be clean, i.e., $T(\cdot)=0$. $\mathcal{D}_{noisy}$ contains the rest unclear examples might be noisy. For $\mathcal{D}_{noisy}$, Promix re-labels them with the most similar known class, thus, we obtain coarse labels $Y^\prime$ for $\mathcal{D}$.

\noindent\textbf{Step 2: Contrastive Learning} We adopt the most popular contrastive learning setups following SupCon \cite{khosla2020supervised} and MoCo \cite{he2020momentum}. Given each example $(\boldsymbol{x}, y)$, we generate its query and key view through randomized data augmentation Aug$(\boldsymbol{x})$, then feed them into the query and key network $g(\cdot),  g^\prime(\cdot)$, yielding a pair of $L_2$-normalized embeddings $\boldsymbol{q} = g(\text{Aug}_q(\boldsymbol{x}))$, $\boldsymbol{k} = g^\prime(\text{Aug}_k(\boldsymbol{x}))$. In implementations, the query network shares the same convolutional blocks as the classifier, followed by a prediction head. Following MoCo, the key network uses a momentum update with the query network. We additionally maintain a queue storing the most current key embeddings $\boldsymbol{k}$ and update it chronologically. To this end, we have the contrastive embedding pool $A = B_q \cup B_k \cup queue$, where $B_q$ and $B_k$ are vectorial embeddings corresponding to the query and key views of the current mini-batch. Given an example $\boldsymbol{x}$, the per-example contrastive loss is defined by contrasting its query embedding with the remainder of the pool $A$:
\begin{equation}
\begin{aligned}
    &\mathcal{L}_{\text {cont}}(g;\boldsymbol{x}, t, A)= \\
    &-\frac{1}{|P(\boldsymbol{x})|} \sum_{\boldsymbol{k}_{+} \in P(\boldsymbol{x})} \log \frac{\exp \left(\boldsymbol{q}^{\top} \boldsymbol{k}_{+} / t\right)}{\sum_{\boldsymbol{k}^{\prime} \in A(\boldsymbol{x})} \exp \left(\boldsymbol{q}^{\top} \boldsymbol{k}^{\prime} / t\right)}. \label{eqa:lcont}
\end{aligned}
\end{equation} 
Here $P(\boldsymbol{x})$ is the set of positive examples for $\boldsymbol{x}$, and $A(\boldsymbol{x}) = A \backslash \{q\}$, $\tau$ is the temperature. Conventionally, the positive examples are from the same class and the negative examples are from different classes. However, in the OSNLL problem, the existence of label noise and open-set examples makes constructing the positive example set $P(\boldsymbol{x})$ particularly challenging. Below, we explain how we perform prototype-based contrastive learning to select positive example set and make use of open-set examples.

\noindent\textbf{Positive Set Selection}
The positive example set is entirely derived from $A(\boldsymbol{x})$, where $A(\boldsymbol{x})$ consists of subsets from $\mathcal{D}_{clean}$ and $\mathcal{D}_{noisy}$, which are referred to as the clean part and the noisy part, respectively. For the clean part, we can directly construct the positive examples based on its label:
\begin{equation}
   P_{clean}(\boldsymbol{x})=\left\{\boldsymbol{k} \mid \boldsymbol{k} \in A(\boldsymbol{x}) \cap \mathcal{D}_{clean}, y={y}^{\prime}\right\},
\end{equation}
where $y^\prime$ and $y$ are the coarse label of example $\boldsymbol{x}$ and $\boldsymbol{k}$.

However, we cannot directly construct the positive example set from the noisy part due to the existence of noise. Therefore, we utilize $\mathcal{D}_{clean}$ to guide the construction of positive example set within the noisy part, by generating an initialized $L_2$-normalized prototype for each class:
\begin{equation}
Q_i = \frac{1}{n_i} \sum_{j=1}^{n_i} q_{ij}, \quad i \in \mathcal{Y},
\end{equation}
where $n_i$ denote the total number of examples for class $i$ in $\mathcal{D}_{clean}$ and $q_{ij}$ represents the $L_2$-normalized embedding of $j$-th example in class $i$. 

Subsequently, we compute the distances between the examples in noisy part and their corresponding class prototypes and then determine whether they belong to the positive example set based on a threshold $\tau$:
\begin{equation}
\begin{aligned}
   P_{noisy}(\boldsymbol{x})=\{\boldsymbol{k} \mid 
   &\boldsymbol{k} \in A(\boldsymbol{x}) \cap \mathcal{D}_{noisy}, y={y}^{\prime},\\ 
   &Distance(\boldsymbol{k}, Q_y) < \tau \},
\end{aligned}
\end{equation}
where $Distance(a, b) = 1 - (a \cdot b) / (\|a\| \|b\|).$
Then, we combine with $P_{clean}$ to obtain a complete positive example set:$P(\boldsymbol{x})=P_{clean}(\boldsymbol{x}) \cup P_{noisy}(\boldsymbol{x}).$
Correspondingly, we can also formalized the incorporated examples from $\mathcal{D}_{noisy}$ as:
$F_q = \mathbb{I}(q \in \mathcal{D}_{noisy} \text{ and } Distance(q, Q_{y^\prime}) < \tau). $
When $F_q = 0$, we treat $q$ as a distinguishable open-set example and refer to this process as the \textbf{open set decision}. When $F_q = 1$, we consider $\boldsymbol{q}$ as a clean example. During the subsequent training process, we update the prototype in a moving-average style:
$ \boldsymbol{Q}_{i}=\operatorname{Normalize}(\gamma \boldsymbol{Q}_{i}+(1-\gamma) \boldsymbol{q}), \text { if } i=y^\prime.$
Here the prototype $Q_i$ of class $i$ is defined by the moving average of the normalized query embeddings $q$ whose coarse label conforms to $i$. $\gamma$ is a tunable hyperparameter.

Also, our classification term can be represented as:
\begin{equation}
\begin{aligned}
\small
    \mathcal{L}_{CLS}= &-\frac{1}{|\mathcal{D}_{clean}|} \sum_{i \in \mathcal{D}_{clean}} \sum_{j=1}^{c} y_{i}^{j} \log \left(p_{i}^{j}\right) - \\
    &\frac{1}{\sum_{k=1}^n\mathbb{I}(F_k=1)} \sum_{i \in \mathcal{D}_{noisy}} \sum_{j=1}^{c} \mathbb{I}(F_i=1) {y^{\prime}}_{i}^{j} \log \left(p_{i}^{j}\right),
\end{aligned}
\end{equation}
where $y_i^j \text{ and } p_i^j$ denote values of the one-hot label and softmax output of the network of example $x_i$ in the $j$-th class.

\noindent\textbf{Training Objective} 
For examples that are distinct from any known classes, their embeddings tend to project toward the inter-class rather than the intra-class. Therefore, by leveraging the nature of these examples, we treat them as delimiters, continuously pushing the known classes away from them, i.e., they are used as negative examples for all known classes. Thus, our contrastive term can be represented as:
\begin{equation}
\begin{aligned}
    \mathcal{L}_{\text {CONT}} &= \frac{1}{|\mathcal{D}_{clean}|} \sum_{i \in \mathcal{D}_{clean}} \mathcal{L}_{\text {cont}}(g;\boldsymbol{x_i}, t, A) + \\
    & \frac{1}{\sum_{i=1}^{n}\mathbb{I}(F_i=1)} \sum_{i \in \mathcal{D}_{noisy}}  \mathbb{I}(F_i=1)  \mathcal{L}_{\text {cont}}(g;\boldsymbol{x_i}, t, A). \label{LCONT}
\end{aligned}
\end{equation}

Finally, we put it all together and jointly train the classifier as well as the contrastive network with the overall loss function as:
\begin{equation}
    \mathcal{L}=\mathcal{L}_{\text {CLS }}+\beta \mathcal{L}_{\text {CONT }},
\end{equation}
in which $\beta$ is a trade-off parameter.

\section{Theoretical Analysis}

% \textbf{Definition 1}\label{cluster} ($\delta$-cluster closeness). Let $S_i$ be the support where $P(u|x_i) > 0$ for any $u \in S_i$. $P(u|x)$ denotes the conditional distribution of an augmented example $u$ given $x$, $S_i$ and $S_j$ are $\delta$-cluster close if $P[u \in S_i \cap S_j|x] \ge \delta$ for any two different examples $x_i$, $x_j$ with $y_i = y_j$.
In this section, we theoretically demonstrate that distinguishable open-set examples can contribute to enhancing the discriminative capabilities of contrastive learning.

\noindent\textbf{Definition 1} (($\sigma$, $\delta$)-Augmentation). The augmentation set $A$ is called a ( $ \sigma $ , $ \delta $ )-augmentation, if for each class $C_k$, there exists a subset $C_k^{0} \subseteq C_k$ (called a main part of $C_k$), such that both $\mathbb{P}[\boldsymbol{x}  \in C_ {k}^ {0} ] \geqslant \sigma \mathbb{P}[\boldsymbol{x} \in C_{k} ]$ where $ \sigma \in  (0,1] $ and $\text{sup}_ {x_ {1}, x_ {2} \in C^{0}_k}, {d_{A}} (x_ {1},x_ {2}) \leqslant  \delta$ hold, where $d_A(\boldsymbol{x}_1,\boldsymbol{x}_2)=\min_{\boldsymbol{x}_1^{\prime}\in A(\boldsymbol{x}_1),\boldsymbol{x}_2^{\prime}\in A(\boldsymbol{x}_2)}\left\|\boldsymbol{x}_1^{\prime}-\boldsymbol{x}_2^{\prime}\right\|$  It represents the augmented distance. \cite{wang2020understanding, khosla2020supervised, duchi2021learning}.

%In other words, the main-part examples locate in a ball with diameter ${\delta}$ (w.r.t. the augmented distance) and its proportion is larger than $\sigma$. Larger $\sigma$ and smaller ${\delta}$ indicate the sharper concentration of augmented data. 
With Definition 1,our analysis will focus on the examples in the main parts with good alignment, i.e., $(C_1^0\cup\cdots\cup C_K^0)\cap S_\varepsilon$, where $S_\varepsilon:=\{\boldsymbol{x}\in\cup_{k=1}^KC_k:\forall\boldsymbol{x}_1,\boldsymbol{x}_2\in A(\boldsymbol{x}),\|f(\boldsymbol{x}_1)-f(\boldsymbol{x}_2)\|\leq\varepsilon\}$ is the set of examples with $\varepsilon$-close representations among augmented data. Furthermore we let $R_{\varepsilon}:=\mathbb{P}\begin{bmatrix}\overline{S_\varepsilon}\end{bmatrix}$.

% The notion of $\delta$-cluster closeness is similar to the cluster assumption in \cite{rigollet2006generalization}. Definition 1 indicates that the augmentations applied to examples should be diverse and extensive, allowing for the potential overlap of any two distinct augmentation distributions within the same class.
 
%With the notion of ($ \sigma $ , $ \delta $)-Augmentation, we investigate the relationship between Eq.(\ref{LCONT}) and cluster properties of representations: the distances between any two clusters. 
We transform Eq.(\ref{LCONT}) to the following formulation:
$$
    \mathcal{L}_{CONT} =  a \sum_{i \in \mathcal{Y}} \sum_{m \in D_i} \mathcal{L}_{align} +
     b  \sum_{\substack{i,j \in \mathcal{Y} \\ i \ne j}} \sum_{\substack{m \in D_{i} \\ t \in D_{j}\\ }} \mathcal{L}_{uniform}, 
$$

\noindent where $D_y$ denotes index set corresponding to true class $y$, $a$ and $b$ are fixed constant related to the example quantity, and 
$\mathcal{L}_{align}=\mathbb{E}_{\substack{u_{m}^{+} \in Pos(u_{m}) \\ }}[\|f(u_{m})-f(u_{m}^{+})\|^{2}]$, $\mathcal{L}_{uniform} = \mathbb{E}_{\substack{u_{t} \in A(u_{m}) \\ u_{t} \notin Pos(u_{m})}}[{-\|f(u_{m})-f(u_{t})\|^{2}}]$, where $u_m \sim P(u|x_m)$, $Pos(\cdot)$ denotes positive example set.

\noindent\textbf{Theorem 1}  \textit{We assume $f$ is normalized by $|f|=r$, and it is $L$-Lipschitz continuity, i.e., for any $\boldsymbol{x}_1,\boldsymbol{x}_2,\|f(\boldsymbol{x}_1)-f(\boldsymbol{x}_2)\|\leq L\|\boldsymbol{x}_1-\boldsymbol{x}_2\|$. We let $p_k:=\mathbb{P}[x\in C_k]\text{ for any }k\in[K].$ Let ${\mu}_{k}=\sum_{i \in D_{k}} \frac{f(u_{i})}{|D_{k}|}$, ${\mu}_{\ell}=\sum_{i \in D_{\ell}} \frac{f(u_{i})}{|D_{\ell}|}$  be centroids of cluster $k$ and cluster $\ell$ with  $k \neq \ell$. If the augmented data is $(\sigma,\delta)$-Augmented, then for any $\varepsilon\geq0$, we have}

$$
\mu_k^\top\mu_\ell \leq \log(\exp\{\frac{\mathcal{L}_\mathrm{uniform}+\tau(\sigma,\delta,\varepsilon,R_\varepsilon)}{p_kp_\ell}\} -\exp(1-\varepsilon)),
$$

\textit{where $\tau(\sigma,\delta,\varepsilon,R_{\varepsilon})$ is a non-negative term, decreasing with smaller $\varepsilon,R_{\varepsilon}$ or sharper concentration of augmented data. The specific formulation of $\tau(\sigma,\delta,\varepsilon,R_\varepsilon)$ and~the~proof~are~deferred~to~the~appendix.}

\noindent\textbf{Remark.} Theorem 1 indicates that the distance between the cluster $k$ and the cluster $\ell$ can be lower bounded by $- \mathcal{L}_{uniform}$. With the introduction of distinguishable open-set examples, a higher lower bound is ensured, thereby enhancing the discriminative nature of different class.

\section{Experiments}
\label{Experiments}
\subsection{Experiment Setting}

\begin{table*}[!ht]
\centering
\small

% \normalsize
% \setlength\tabcolsep{10pt}
% \resizebox{\textwidth}{!}
{
    \begin{tabular}{lccc|ccc}
    \toprule[1.5pt]
    \multirow{2}{*}{Methods} & \multicolumn{3}{c}{CIFAR80N} & \multicolumn{3}{c}{CIFAR100N}  \\ 
    \cline{2-7} &  \textit{Sym - 20\%} & \textit{Sym - 80\%} & \textit{Asym - 40\%} & \textit{Sym - 20\%} & \textit{Sym - 80\%} & \textit{Asym - 40\%} \\ 
    \midrule[1pt]
    Standard          & 29.37 $\pm$ 0.09 & 4.20 $\pm$ 0.07   & 22.25 $\pm$ 0.08 & 35.14 $\pm$ 0.44 & 4.41 $\pm$ 0.14 & 27.29 $\pm$ 0.25  \\
    Decoupling                & 43.49 $\pm$ 0.39 & 10.01 $\pm$ 0.29  & 33.74 $\pm$ 0.26 & 33.10 $\pm$ 0.12 & 3.89 $\pm$ 0.16 & 26.11 $\pm$ 0.39  \\
    Co-teaching                 & 60.38 $\pm$ 0.22 & 16.59 $\pm$ 0.27 & 42.42 $\pm$ 0.30 & 43.73 $\pm$ 0.16 & 15.15 $\pm$ 0.46  & 28.35 $\pm$ 0.25  \\
    Co-teaching+               & 53.97 $\pm$ 0.26 & 12.29 $\pm$ 0.09 & 43.01 $\pm$ 0.59 & 49.27 $\pm$ 0.03 & 13.44 $\pm$ 0.37 & 33.62 $\pm$ 0.39  \\
    JoCoR                      & 59.99 $\pm$ 0.13 & 12.85 $\pm$ 0.05 & 39.37 $\pm$ 0.16 & 53.01 $\pm$ 0.04 & 15.49 $\pm$ 0.98 & 32.70 $\pm$ 0.35  \\
    MoPro                   & 65.60 $\pm$ 0.34 & 30.29 $\pm$ 0.21 & 60.22 $\pm$ 0.12 & 54.22 $\pm$ 0.26 & 28.32 $\pm$ 0.34 & 49.69 $\pm$ 0.45  \\
    NGC                        & 74.26 $\pm$ 0.23 & 36.36 $\pm$ 0.48 & 65.73 $\pm$ 0.44 & 68.47 $\pm$ 0.28 & \textbf{37.17} $\pm$ 0.41 & 64.79 $\pm$ 0.38  \\
    Jo-SRC                        & 65.83 $\pm$ 0.13 & 29.76 $\pm$ 0.09 & 53.03 $\pm$ 0.25 & 58.15 $\pm$ 0.14 & 23.80 $\pm$ 0.05 & 38.52 $\pm$ 0.20  \\
    PNP           & 67.00 $\pm$ 0.18 & 34.36 $\pm$ 0.18 & 61.23 $\pm$ 0.17 & 64.25 $\pm$ 0.12 & 31.32 $\pm$ 0.19 & 60.25 $\pm$ 0.21 \\ 
    \midrule[1pt]
    \textbf{CECL}      & \textbf{77.23} $\pm$ 0.26 & \textbf{37.21} $\pm$ 0.11 & \textbf{68.48} $\pm$ 0.14 &\textbf{69.20} $\pm$ 0.09 & 36.37  $\pm$ 0.12 & \textbf{65.49}  $\pm$ 0.24  \\ 
    \bottomrule[1.5pt]
    \end{tabular}
}
% \captionsetup{font=small}
\caption{Test accuracy (\%) comparison on synthetic noisy datasets CIFAR80N and CIFAR100N. The average mean and standard deviation results over the last 10 epochs are recorded. Bold values represent the best methods.}
\label{table1}
\end{table*}

\begin{table*}[!ht]
\centering
\small
% \normalsize
\setlength\tabcolsep{10pt}
% \resizebox{\textwidth}{!}
{
\begin{tabular}{lcccc}
\toprule[1.5pt]
Methods  & Web-Aircraft & Web-Bird & Web-Car & Food101N \\ \midrule[1pt]
Standard &  60.80 & 64.40 & 60.60 & 84.51    \\
Decoupling \cite{malach2017decoupling}  & 75.91 & 71.61 & 79.41   & -        \\
Co-teaching \cite{han2018co}            & 79.54 & 76.68 & 84.95   & -        \\
CleanNet-hard \cite{lee2018cleannet}       & -     & -     & -       & 83.47    \\
CleanNet-soft \cite{lee2018cleannet}      & -     & -     & -       & 83.95    \\
SELFEI \cite{song2019selfie}              & 79.27 & 77.20 & 82.90   & -        \\
PENCIL \cite{yi2019probabilistic}        & 78.82 & 75.09 & 81.68   & -        \\
Co-teaching+ \cite{yu2019does}            & 74.80 & 70.12 & 76.77   & -        \\
Deep-Self \cite{han2019deep}               & -     & -     & -       & 85.11    \\
JoCoR \cite{wei2020combating}             & 80.11 & 79.19 & 85.10   & -        \\
AFM \cite{peng2020suppressing}             & 81.04 & 76.35 & 83.48   & -        \\
CRSSC \cite{sun2020crssc}                & 82.51 & 81.31 & 87.68   & -        \\
Self-adaptive \cite{huang2020self}      & 77.92 & 78.49 & 78.19   & -        \\
DivideMix \cite{li2020dividemix}         & 82.48 & 74.40 & 84.27   & -        \\
Jo-SRC \cite{yao2021jo}                    & 82.73 & 81.22 & 88.13   & 86.66    \\
Sel-CL \cite{li2022selective}              & 86.79 & 83.61 & 90.40   & -    \\
PLC \cite{zhang2021learning}               & 79.24 & 76.22 & 81.87   & -        \\
NGC \cite{wu2021ngc}                       & 85.94 & 83.12 & 91.83   & 89.64        \\
Peer-learning \cite{sun2021webly}        & 78.64 & 75.37 & 82.48   & -        \\
PNP \cite{sun2022pnp}                    & 85.54 & 81.93 & 90.11   & 87.50    \\ \midrule[1pt]     
\textbf{CECL}                           & \textbf{87.46}  & \textbf{83.87} & \textbf{92.61}  & \textbf{90.24}    \\ \bottomrule[1.5pt]
\end{tabular}
}
% \captionsetup{font=small}
\caption{Test accuracy(\%) comparison on four real-world noisy datasets. Bold values represent the best methods.}
\label{table2}

\end{table*}
% \textbf{Datasets} Following exactly the same setting as \cite{sun2022pnp}, we evaluate our CECL approach on two synthetic datasets  CIFAR80N and CIFAR100N and four real-world datasets Web-Aircraft, Web-Bird, Web-Car and Food101N. CIFAR80N and CIFAR100N stem from CIFAR100 \cite{krizhevsky2009learning}. Specifically, CIFAR80N is an open-set noisy dataset with $20$ unknown classes, and CIFAR100N is a closed-set noisy dataset to test the generality of methods to deal with both closed-set and open-set label noise. We adopt two classic noise structures symmetric and asymmetric, with three different noise ratios: \emph{Sym-20\%}, \emph{Sym-80\%} and \emph{Asym-40\%}. Web-Aircraft, Web-Bird, and Web-Car are three real-world web image datasets for fine-grained vision categorization. Within each dataset, more than $25\%$ of the training examples are associated with unknown (asymmetric) noisy labels. Plus, these datasets do not provide any label verification information, making them more practical and better represent the ONSLL problem. Food101N \cite{lee2018cleannet} is another large-scale real-world noisy dataset, containing 101 different food categories over 310k training examples. And its noise ratio hasn't been calculated, making it hard to specialize in a specific noise type. 

\textbf{Datasets} We evaluate our CECL approach using the setup from \cite{sun2022pnp} on several datasets. CIFAR80N and CIFAR100N, generated from CIFAR100 \cite{krizhevsky2009learning}, offer distinct challenges. CIFAR80N contains an open-set noisy dataset with $20$ unknown classes, while CIFAR100N is a closed-set noisy dataset examining methods' versatility with different label noise scenarios (\emph{Sym-20\%}, \emph{Sym-80\%}, \emph{Asym-40\%}). Real-world datasets (Web-Aircraft, Web-Bird, Web-Car) tackle fine-grained vision categorization, presenting over $25\%$ training examples with ambiguous noisy labels and lacking label verification. Additionally, Food101N \cite{lee2018cleannet} showcases 101 food categories with more than 310k examples, posing a challenge due to its unspecified noise ratio and diverse noise types.

\noindent\textbf{Comparison methods} On CIFAR80N and CIFAR100N, referring \cite{sun2022pnp}, we compare with the following baselines: \textbf{Standard} which conducts cross entropy loss minimization, \textbf{Decoupling} \cite{malach2017decoupling}, \textbf{Co-teaching} \cite{han2018co}, \textbf{Co-teaching+} \cite{yu2019does}, \textbf{JoCoR}, which are commonly used denoise approaches in learning with noisy labels, and recently proposed methods \textbf{MoPro}, \textbf{NGC}, \textbf{Jo-Src}, \textbf{PNP} customized for OSNLL. On the four real-word datasets, we compare with \textbf{Standard}, \textbf{NGC}, \textbf{Jo-Src}, \textbf{PNP} and \textbf{16} other closed-set noisy label learning approaches as shown in Table \ref{table2}. 

 For baselines, their experimental results are directly adopted from \cite{sun2022pnp}. For our CECL method, we maintain the same backbone as PNP. % on CIFAR80N and CIFAR100N, we employ the ResNet18 \cite{he2016deep} as the feature extractor, on Real-world datasets(Web-Aircraft, Web-Bird, Web-Car and Food101N), we employ the pretrained ResNet50 as the feature extractor, the classifiers and MLP for all datasets are 1-layer linear classifiers and 3-layer neural networks, respectively. 
For optimization, we uniformly adopt SGD with a momentum of $0.9$, a learning rate decay strategy of CosineAnnealingLR, and a batch size of 256, 16, 96 for CIFAR, Web, Food101N datasets.
% (参考：1
%  %For RSLR, on Digits tasks, we employ the CNN structure~\cite{ganin2015unsupervised} as the feature extractor, on Office-31 and Office-Home, we use the ResNet-50 pre-trained on ImageNet as a feature extractor. The classifiers for LNets and TNet on all datasets use 3-layer neural networks. We train the model by using mini-batch SGD with a momentum $0.9$, a weight decay of $0.0004$, and a batch size of 32 (128 in Digits tasks). We set the learning rate of $G$ and $C$ respectively as $0.01$ and $0.1$. The initial candidate factor of pseudo-labels samples $\gamma_{0} $ is set as $0.4$ on Office-31/Office-Home and $0.1$ on Digits. Before starting training, we warm up the model on noisy data with $\mathit{E }_{warm} $ epochs according to the noise rate ratio. We set the clean probability threshold $\tau$ = 0.6. For data augmentation, we follow FixMatch~\cite{25} and set the number of augmentations as $2$.

%  参考2：
%  %The epochs of the pretraining and the stage $1,2$ in KD-Crowd are $10,  30, 10$. For the real-world datasets, we use ShuffleNet V2\cite{shufflenetV2} on LabelMe as backbone trained for $75$ epochs for baselines. The epochs of the pretraining and the stage $1,2$ in KD-Crowd are $5, 60, 10$. 
%  )

\subsection{Classification Comparison}
Table \ref{table1} and Table \ref{table2} respectively show results for the synthesized and real-world datasets. Regarding the error bars in Table \ref{table1}, we have maintained the same setting as PNP, and reported the average and standard deviation of the last 10 epochs, under the condition of a random seed value 0. Bold values represent the best results among all methods.

Table \ref{table1} shows results for the $20\%, 80\%$ symmetric noise case and $40\%$ asymmetric noise case. By varying the structure and ratio of label noise, we try to provide a more comprehensive view of our proposed method. %It shows that our method consistently outperforms state-of-the-art approaches across different noise scenarios. 
Compared with existing OSNLL methods which minimize the impact of open-set examples, CECL wisely leverages them to learn better representation space for the known classes, leading to significant performance gains. Surprisingly, on the closed-set label noise dataset CIFAR100N, our proposed CECL still achieves better performance than the closed-set baselines. We hypothesize that this may be due to the models' tendency to treat the difficult and ambiguous examples as open-set examples, which is more reasonable than forcing them to some incorrect classes and resulting in negative effects. %avoid the negative rather than may aid in the learning process. %The reason behind this is that model's performance at first stage is much better in less inferior cases and the class expansion contrastive learning scheme partly relies on the results retained in the first step. It is also worth mentioning that our method also asks for much fewer hyper-parameter tuning efforts and regularization terms compared with other methods like PNP and Jo-SRC.

Table \ref{table2} displays the experimental results on the four real-world noisy datasets. Compared with the synthetic datasets, the types of label noise in these datasets become more complicated with relatively lower noise ratios. We can observe similar superiority of our CECL method, which outperforms other methods. Such results validate our method works robustly on complex scenarios in learning with noise.

\subsection{Ablation Study}
\begin{table}[t]
\centering
\small
% \normalsize
\setlength\tabcolsep{1.5pt}
\resizebox{\columnwidth}{!}
{
\begin{tabular}{l|cc|cc}
\toprule[1.5pt]
              & \textit{CONT}  & \textit{OSD}    & \textit{sym 20\%} & \textit{asym 40\%} \\
\midrule[1pt] 
CECL \textit{w/o} \textit{CONT} \& \textit{OSD} & \ding{56}    & \ding{56}    &   70.54       & 62.41 \\
CECL \textit{w/o} \textit{CONT}   & \ding{56}    &  \ding{52}     &   73.17      & 64.57  \\
CECL \textit{w/o} \textit{OSD}    & \ding{52}    & \ding{56}    &    72.87      & 64.26  \\
CECL \textit{with} \textit{RDOS}     & \ding{52}    &  \ding{52}     & 75.86         & 66.73 \\
CECL                                   & \ding{52}    &   \ding{52}        &   \textbf{77.23}       & \textbf{68.48}  \\
\bottomrule[1.5pt]
\end{tabular}
}
% \captionsetup{font=footnotesize}

\caption{Ablation study on CIFAR80N with \textit{sym 20\%} and \textit{asym 40\%}, \textit{OSD} denotes \textit{open-set decision}, \textit{RDOS} denotes \textit{removing distinguishable open-set examples}.}
\label{table3}
\end{table}

In this subsection, we present our ablation results to show the effectiveness of CECL. We ablate two key components of CECL: contrastive learning (\textit{CONT} ) and open-set decision (\textit{OSD}). %, showing that proper detection and usage of open-set examples plays a crucial role in OSNLL. %\textit{OSD} means determining which examples will be considered as distinguishable open-set. 

%five variants to show that proper detection and usage of open-set examples play a crucial role in OSNLL: 1) which 

%each component of CECL:%two key components:  the contrastive learning ($\mathcal{L}_{CONT}$) and open-set decision (\textit{OSD}), \textit{OSD} means determining which examples will be considered as distinguishable open-set. 

We compare CECL from four views: 1) To validate the effectiveness of Step 2: \textit{CECL w/o CONT \& OSD}, which uses neither \textit{CONT} nor \textit{OSD}, that is, our Step 1, Promix; 2) To validate the effectiveness of contrastive component: \textit{CECL w/o CONT}, which removes the contrastive learning and only trains a classifier with clean examples and closed-set examples with corrected label identified by the prototype-based open-set decision; 3) To validate the effectiveness of \textit{OSD}: \textit{CECL w/o OSD}, specifically, we regard all the examples identified as noise in Step 1 as known classes examples and let them participate in contrastive learning, that is, the conventional supervised contrastive learning; 4) To validate the benefits brought by the distinguishable open-set examples: \textit{CECL with removing distinguishable open-set examples (RDOS)}, this means that we ignore the distinguishable open-set examples. As shown in Table \ref{table3},  the combination of \textit{CONT} and \textit{OSD} can achieve the best performance.

% From Table \ref{table3}, we can observe: 1) Step 1 overlooks the consideration of distinguishable open-set examples, thus yielding unsatisfactory results, which demonstrate the effectiveness of Step 2; 2) The lack of contrastive learning can affect the model's representational capacity, thereby influencing the performance; 3) Simply treating all open-set examples as closed-set ones leads to suboptimal performance; 4) introducing distinguishable open-set examples into contrastive loss can improve the performance. In sum, as shown in Figure \ref{delimiter}, at the representation level, utilizing \textit{OSD} to select appropriate examples for contrastive learning can lead to the expansion of class boundaries and enhance the differentiation of known classes. Moreover, representation learning significantly aids in correcting noise. Therefore, the combination of $\mathcal{L}_{CONT}$ and \textit{OSD} can achieve the best performance.

\subsection{Further Analysis}
\begin{figure}[htbp]
  \centering
  \subfigure[PNP]{\includegraphics[width=0.45\columnwidth]{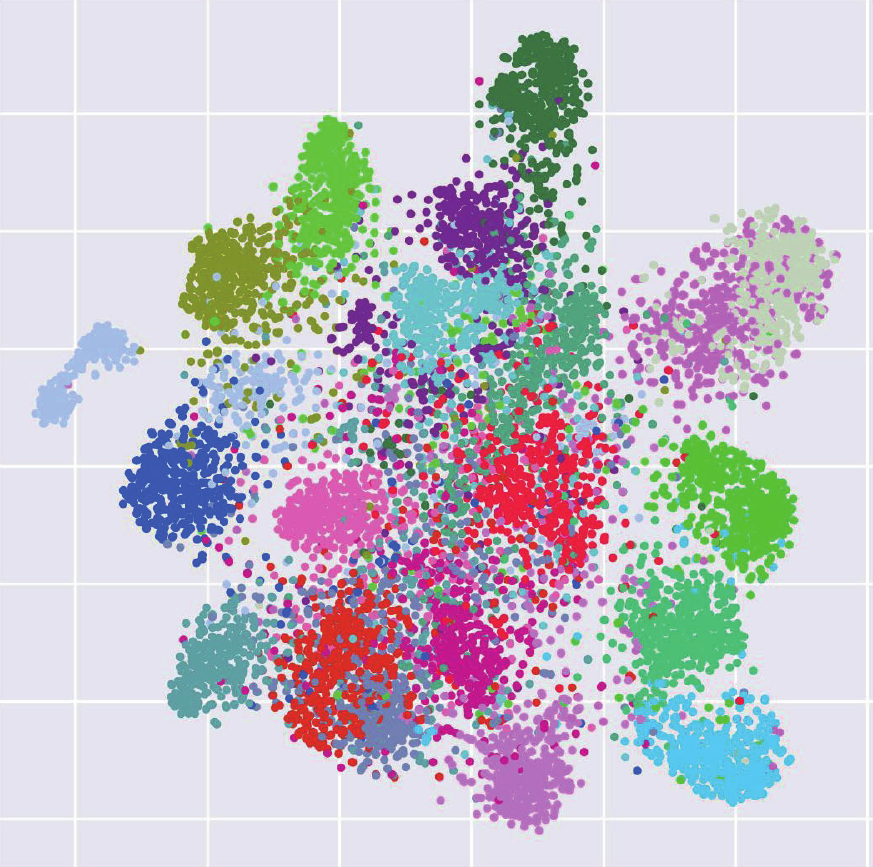}}
  \hfill
  \subfigure[CECL]{\includegraphics[width=0.45\columnwidth]{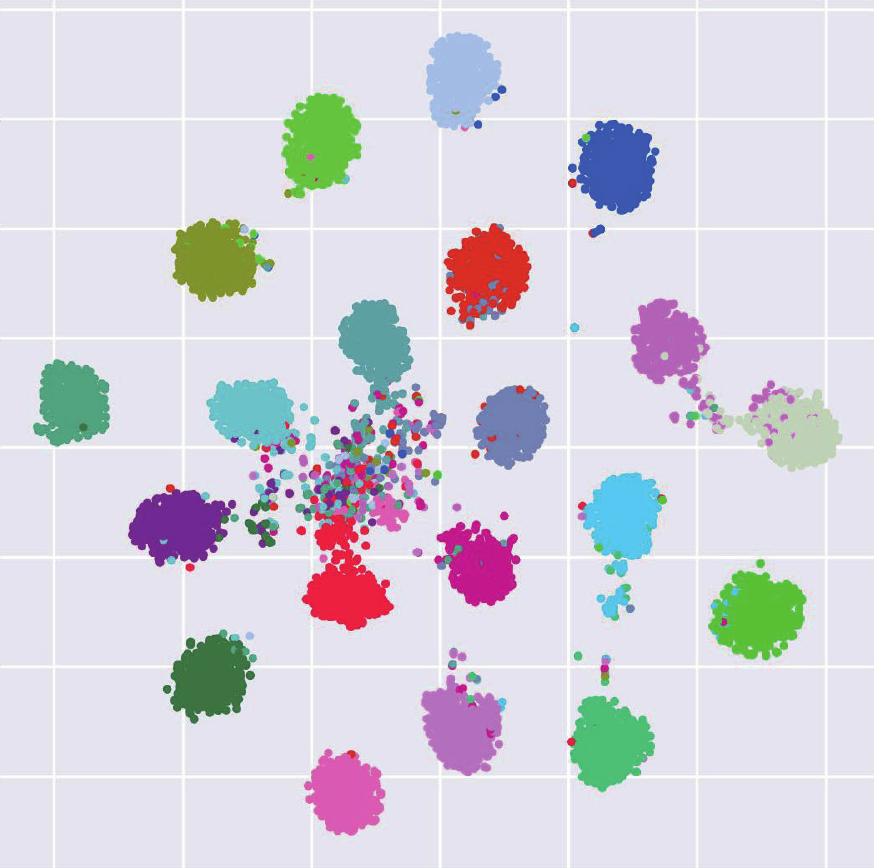}}

  % \subfigure[PNP]{\includegraphics[width=0.45\columnwidth]{CameraReady/LaTeX/Figure/tsne_3.pdf}}
  % \hfill
  % \subfigure[CECL]{\includegraphics[width=0.45\columnwidth]{CameraReady/LaTeX/Figure/tsne_4.pdf}}

  \caption{The t-SNE on CIFAR80N with \textit{Sym 20\%.}} %(top row), \textit{Asym 40\%} (bottom row).}
  \label{t-sne}
\end{figure}

\noindent\textbf{Feature Space Visualization}
The CECL framework consistently emphasizes improving representation learning. To evaluate its progress in representation capabilities, we compared it with PNP and visualized the results using t-SNE\cite{van2008t-SNE}, as shown in Figure \ref{t-sne}.

% \begin{figure}[htbp] 
% \begin{center}
% \centerline{\includegraphics[width=0.8\columnwidth,height=2.3in]{Figure/tsne2.eps}}
% \caption{The t-SNE on CIFAR80N with \textit{Sym 20\%} (top row), \textit{Asym 40\%} (bottom row).}
% \label{t-sne}
% \end{center}
% \end{figure}

We show the t-SNE of the randomly selected $20$ classes on CIFAR80N with symmetric $20\%$ noise. The features learned by our method are more discriminative in every class compared with PNP, which verifies our method can learn better features under different noise levels and types. %(top row) and asymmetric $40\%$ noise (bottom row)

\begin{figure}[htbp]
\centering

\centerline{\includegraphics[width=0.8\columnwidth, height=5cm]{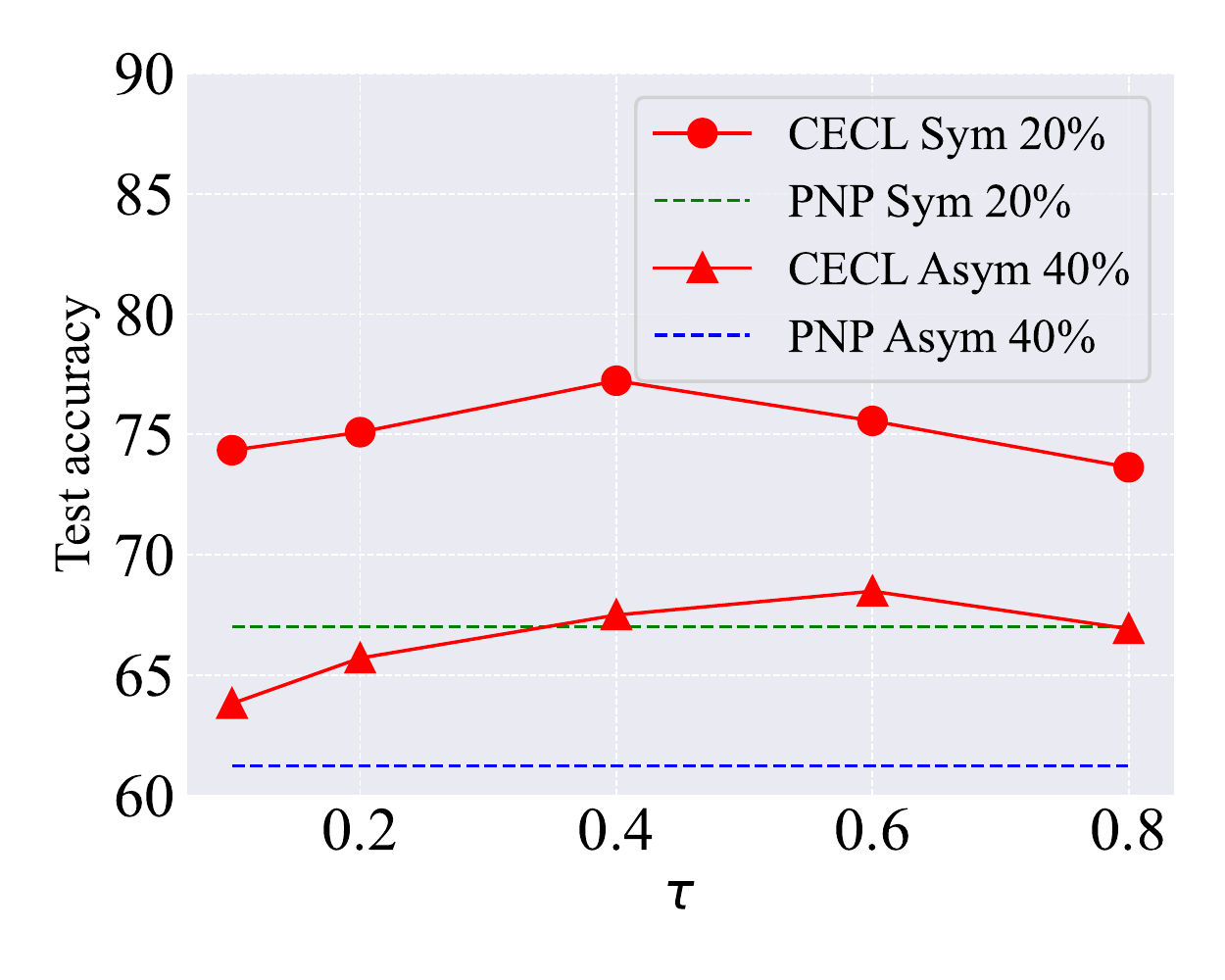}}
\caption{Sensitivity analysis of hyper-parameter $\tau$ on CIFAR80N with \textit{Sym 20\%} and \textit{Asym 40\%}.}
\label{sensitivity}
\end{figure}

\noindent\textbf{Parameter Sensitivity}
%Our method's sensitivity primarily lies in the threshold $\tau$ for open-set decisions, which is critical for identifying distinguishable open-set examples, constructing positive examples, and enabling the reuse of open-set examples. 
We explore the impact of varying $\tau$ on the performance of our method. We conduct experiments on the CIFAR80N with \textit{Sym20\%} and \textit{Asym40\%}, and the results are shown in Figure \ref{sensitivity}. The results indicate that the performance significantly higher than PNP at different values of $\tau$, demonstrating the robustness of our method.

\section{Conclusion}
This paper tackles the task of learning from real-world open-set noisy labels. Unlike conventional methods that merely identify and minimize the impact of open-set examples, we reveal the Class Expansion phenomenon, demonstrating their ability to enhance learning for known classes. Our proposed CECL approach integrates appropriate open-set examples into known classes, treating the remaining distinguishable open-set examples as delimiters between known classes. This strategy, coupled with a modified contrastive learning scheme, boosts the model's representation learning.

% This paper addresses the challenge of learning from real-world open-set noisy labels. While traditional approaches strictly identify open-set examples and minimize their impact, we have shown the Class Expansion phenomenon of open-set examples and their power to boost known class learning.  We propose the CECL approach, which incorporates appropriate open-set examples into the known classes and treats the remaining distinguishable open-set examples as delimiters between known classes, and enhances the model's representation learning through a modified contrastive learning scheme. For the first time in OSNLL, our CECL approach embraces open-set examples instead of excluding and treating them as a nuisance. We acknowledge that such practice may introduce interference to the model's open-set recognition ability. Thus, in future work, we will further refine the proposed framework to address this issue.

\section{Acknowledgments}
This work was supported by the National Key R\&D Program of China (2022ZD0114801), National Natural Science Foundation of China (61906089), Jiangsu Province Basic Research Program (BK20190408).

\bibliography{main}

\end{document}